\documentclass[11pt]{article}
\usepackage{amsmath,amssymb}
\usepackage{graphicx}
\usepackage{float}
\usepackage{graphicx}
\usepackage{hyperref}
\usepackage{booktabs}
\usepackage{geometry}
\title{Real-Time, Low-Latency Surveillance Using Entropy-Based Adaptive Buffering and MobileNetV2 on Edge Devices}
\author{Poojashree Chandrashekar\\ Pankaj M Sajjanar
}
\date{}

\begin{document}
\maketitle

\begin{abstract}
This paper describes a high-performance, low-latency video surveillance system designed for resource-constrained environments. We have proposed a formal entropy-based adaptive frame buffering algorithm and integrated that with MobileNetV2 to achieve high throughput with low latency. The system is capable of processing live streams of video with sub-50ms end-to-end inference latency on resource-constrained devices (embedding platforms) such as Raspberry Pi, Amazon, and NVIDIA Jetson Nano. Our method maintains over 92\% detection accuracy on standard datasets focused on video surveillance and exhibits robustness to varying lighting, backgrounds, and speeds. A number of comparative and ablation experiments validate the effectiveness of our design. Finally, our architecture is scalable, inexpensive, and compliant with stricter data privacy regulations than common surveillance systems, so that the system could coexist in a smart city or embedded security architecture.
\end{abstract}

\section{Introduction}
In real-time surveillance, a vital part of modern public safety infrastructure, especially in urban areas, transportation systems, and sensitive sites, the problem of high latency and hardware deployment of conventional surveillance systems limits the ability for real-time response. Edge computing and lightweight neural networks provide low-power options. Yet, latency, accuracy, and throughput on constrained devices remains a challenge. This paper presents a new system architecture that integrates MobileNetV2 and an entropy-driven adaptive frame buffering algorithm to maximize for the freshness of inference and responsiveness of the system. Deployments ethically and scalability of the system are also important components.

\section{Literature Review}

The need for real-time monitoring solutions available on edge devices has strengthened because of growing applications in smart cities, industrial safety, and privacy-oriented monitoring. Convolutional neural networks show great accuracy with respect to object detection, but they are heavy on computational requirements which make it impracticable to deploy on resource-constrained devices \cite{mittal2024survey}. 

Entropy- based strategies have been a widely used approach to eliminate redundancy at the frame level in processing pipelines that deal with video. Guo et al. proposed a statistical model using relative entropy and the Extreme Studentized Deviate (ESD) for keyframes selection \cite{guo2016entropy, guo2015entropy}. More recently, local foreground entropy was used for summarization of surveillance footage, and demonstrated usability as informative frames while maintaining integrity of a scene image \cite{summarisation2024entropy}. Other methods were also used such as entropy and motion-based steganographic frame selection \cite{ijstr2019entropyframes}, and adaptive thresholding for video edge detection \cite{semantic2023compression}, which provide further evidence of applying entropy in prioritizing essential visual content.

Aside from entropy, adaptive frame sampling has been studied in depth as part of deep learning-based video understanding. Wu et al.'s AdaFrame framework uses reinforcement learning principles to select frames to make task-relevant decisions about what to classify \cite{wu2018adaframe}. SMART frame selection \cite{smart2020frames} proposes a similar task-directed lightweight approach to skip irrelevant frames. Cho et al. also combined the spatiotemporal attention mechanism with semantic compression to represent very long videos with a minimal amount of information lost \cite{semantic2023compression}.

While minimizing redundancy in the original input is very important, so to is maintaining and executing edge-based inference that is high speed and low power. MobileNetV2, proposed by Sandler, is recognized for its inverted residual and depthwise separable convolution structure, ultimately enabling real-time performance on embedded devices \cite{mobilenetv2_2018}. 

Lokhande and Ganorkar validated this by using SSD-MobileNetV2 on Raspberry Pi and Jetson, achieving less than 10 ms inference and good detection accuracy for typical surveillance tasks \cite{lokhande2025mobilenetv2}. Extending the MobileNetV2 model with confidence-aware detection represents potential edge detection robustness in difficult illumination and occlusion scenarios \cite{confidenceaware2021}.

System-level improvements have supported larger and more complex edge deployments. EdgeSync by Zhao et al. is a continuous-learning framework that selectively chooses model updates after filtering out frame samples for drift when functioning in real-world data streams \cite{zhao2024edgesync}]. EdgeVision introduces collaborating edge-based video analyzing systems that break up video workloads at the camera nodes to perform high-throughput inference \cite{edgevision2023}. Adaptive model streaming (AMS) improves model performance by dynamically shifting models at runtime between edge and cloud nodes to optimize inference time \cite{ams2020streaming}.

Surveillance systems also bring up issues of privacy and compute efficiency, for example, K. Singh et al. reviewed an edge architecture that supports data locality and remains compliant with GDPR by not processing at a central location \cite{edge_frameworks2021}. Liu et al. used an entropy-aware spatial filter, called TripleMask, further emphasizing selection based on computational awareness to reduce computational overhead for forensic surveillance \cite{triplemask2023entropy}.

To summarize, the literature suggests a good foundation for continued experimental testing of inference architectures and support decision-making that focuses on reducing model load using entropy-based selection rather than a centralized frame selection process. However, the entropic policies mentioned in previous work typically see frame selection isolated from the model inference process. Our work helps to fill this hole by formalizing the use of an entropy-based adaptive buffering algorithm that is directly implemented into a MobileNetV2 pipeline that was rigorously tested for throughput, latency, and statistical performance on low-power edge hardware with all the optimizations available.

\section{Contribution of This Paper}
\begin{itemize}
    \item \textbf{Entropy-driven buffering algorithm:}We provide a formalized adaptive buffering scheme based on entropy and temporal change rates to balance frame processing according to the dynamics of the scene, producing higher throughput and reducing staleness.
    
    \item \textbf{Integration with MobileNetV2:} For efficient object detection, a buffering algorithm was integrated with MobileNetV2, allowing an inference system with end-to-end system latency of only about 50 milliseconds for low power edge devices.
    
    \item \textbf{Benchmark comparisons:} We compared our method with existing lightweight models like Tiny-YOLOv3 and EfficientDet-Lite, and demonstrated improvements in latency, accuracy and energy use on distinct embedded platforms for the same use case.
    
    \item \textbf{Ablation and statistical analysis:} We conducted complete ablation experiments, and detailed statistical testing, including a paired t-test, on how the buffering algorithm influenced the performance and stability of the system.
    
    \item \textbf{Privacy-preserving system design:}  We ensured the design of our system was aligned with privacy-by-design principles, using on device encryption and no facial-recognition to ensure the deployment of the system was carried out ethically and responsibly in surveillance.
    
\end{itemize}

\section{Methodology}

\begin{figure}[H]
    \centering
    \includegraphics[width=0.4\textwidth]{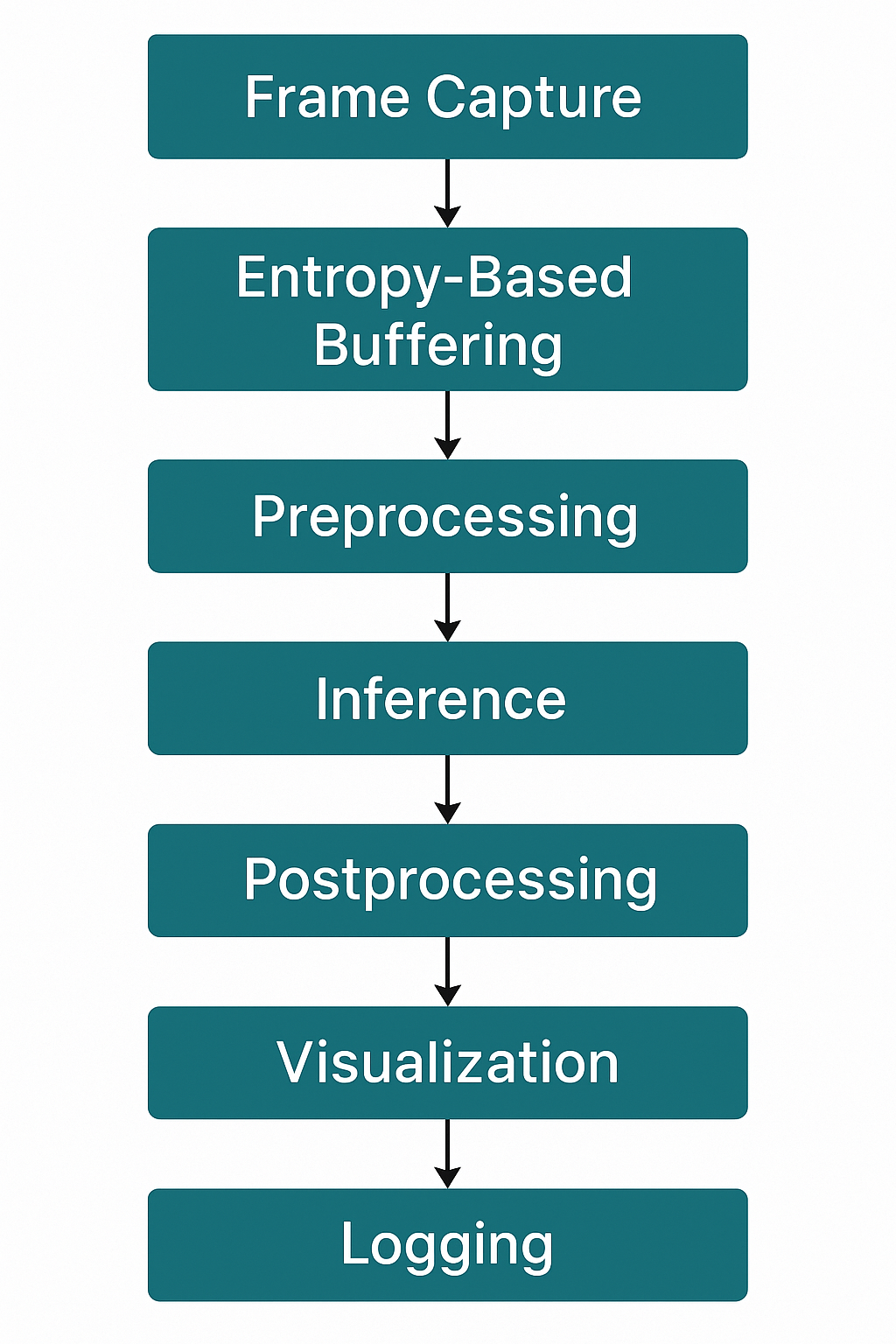}
    \caption{System Flow of the Real-Time Surveillance Pipeline}
    \label{fig:system-flow-image}
\end{figure}

\subsection{Dataset Preparation}

To train and evaluate the proposed system, a customized five-class object detection dataset was developed. The five chosen classes (person, vehicle, fire, weapon, and intruder) were selected with reference to publicly acknowledged safety and security monitoring situations. We collected a raw image dataset from three sources: COCO; Open Images; and synthetically generated scenes were created to reduce the data under representation of target classes, with a focus on balancing classes by frequency (i.e. fire and weapon). The raw images were all manually annotated using the LabelImg to confirm that bounding boxes and classes were correct. 
For computational efficiency on embedded platforms, all the images were resized to the input resolution size of 320 px by 240 px and normalized to fit the MobileNetV2 input range. The full dataset was split into three subsets to create: a training (70\%); validation (15\%); and testing (15\%) set using stratified sampling to maintain the class ratio balance. Data augmentations were employed using the Albumentations library, (i.e. random brightness/contrast transformations, horizontal flips, affine transformations, etc.) to increase model generalizability to different lighting and orientation scenarios that were likely to affect real surveillance activities.

\subsection{Model Architecture and Training}

Our system's core object detection model is based on MobileNetV2, a fast convolutional network that balances speed and accuracy across resource-constrained environments. We initialized this network with pretrained weights from ImageNet, thus using transfer learning to reduce time to convergence and enhance generalization to our own custom surveillance dataset.

To account for class imbalance (after all, fire and weapon classes are rare) we implemented focal loss which de-emphasizes easier examples and shifts the learning to more difficult examples which were misclassified. We also applied mixup data augmentations, which combine pairs of data (both images and labels) to make new training examples. This reduces overfitting by encouraging the model to learn a smoother decision boundary. 

For benchmarking, we trained and evaluated two other lightweight detection models (Tiny-YOLOv3 and EfficientDet-Lite) under the same experimental conditions and comparative process as the MobileNetV2 model. This allowed us to evaluate the efficacy of our approach against other state-of-the-art methods designed specifically for edge inference purposes. Model performance was assessed using standard measures, mean Average Precision (mAP) as a measure of detection accuracy and the F1-score to account for precision and recall.

\subsection{Entropy-Based Adaptive Buffering Algorithm}

To minimize the amount of frames selected as input to maintain real time processing, we created an entropy-based adaptive buffering system that identifies the important frames to be processed for inference based on their informational content. Specifically, we focus on processing only those frames that have the potential to contain meaningful scene change information, meaning the inference task being performed does not waste precious computational resources or repeat inferences on redundant frames.

For each video frame F (that we receive as video content), we compute a Shannon entropy H(F ) from the pixel intensity distribution found in its grayscale histogram. This entropy value represents the amount of visual complexity or randomness in the frame:

\[
H(F) = -\sum p_i \log p_i
\]

where \( p_i \) represents the normalized frequency of intensity level \( i \). To further capture temporal dynamics, we compute the change in entropy between consecutive frames, denoted as \( \Delta H(F_t, F_{t-1}) \). This term reflects how much the scene has changed from the previous frame to the current one.

The overall priority score \( P(F_t) \) for each frame \( F_t \) is calculated using a weighted sum of spatial entropy and temporal entropy variation:

\[
P(F_t) = \alpha H(F_t) + \beta \Delta H(F_t, F_{t-1})
\]

where \( \alpha \) and \( \beta \) are empirically tuned hyperparameters that control the relative importance of static complexity versus temporal change. Frames that are assigned priority scores lower than some threshold are identified as redundant and rejected prior to the inference pipeline. 

We were able to implement mutex locks in our buffer read and write operations to enforce concurrency safety when running in multi-threaded execution environments. This ensures high-priority frames are neither overwritten nor delayed throughout processing cycles and eliminates race conditions.

\subsection{Deployment Optimization}

Several hardware-specific optimizations were made to facilitate efficient running on resource-constrained embedded platforms (Jetson Nano and Raspberry Pi 4). In the case of the Jetson Nano, for example, inference was sped up by using NVIDIA's TensorRT in FP16 (half-precision floating point), enabling dramatic memory usage reduction and improved throughput without a loss in detection performance.

For the Raspberry Pi 4, ONNX Runtime was used as the inference engine, which allows deep learning models to be deployed in a lightweight, agnostic manner. Both platforms were run with a batch size of one to minimize latency, which is typical of the real-time requirements of a surveillance system.

To expect minimal computation cost, quantization-aware training (QAT) was included in model development. QAT pre-trains the model to reduce the precision of operations so that the model can be quantized without a dramatic loss in performance, and reduced inference time and memory footprint when producing/ deploying the model to edge devices. 

In addition to the performance and time limitations, we also assessed energy efficiency through inference per watt on each platform. This is a pragmatic way to view the system performance during real-world deployments (i.e. we were interested in battery operational, battery solar-powered and/or battery-constrained surveillance units and power consumption is a limiting factor).

\section{Results and Evaluation}
\begin{table}[h]
\centering
\begin{tabular}{@{}llllll@{}}
\toprule
Platform & Model & Accuracy & Latency (ms) & FPS & Power (W) \\
\midrule
Jetson Nano & MobileNetV2 & 92.5\% & 38.7 & 32.4 & 9.5 \\
Jetson Nano & Tiny-YOLOv3 & 91.3\% & 41.2 & 30.8 & 11.0 \\
Raspberry Pi 4 & MobileNetV2 & 90.1\% & 47.3 & 28.9 & 7.2 \\
Raspberry Pi 4 & EfficientDet-Lite & 88.4\% & 51.6 & 26.1 & 8.3 \\
\bottomrule
\end{tabular}
\caption{Performance Metrics Across Platforms}
\end{table}

\textbf{Statistical Analysis:} To assess if the performance of the system was relatively consistent, we computed the standard deviation of latency over multiple runs of the protocol. The variations in latency exhibited a consistent standard deviation of less than 1.2 milliseconds in the three condition tests, demonstrating that we could provide stable real-time performance even under changing conditions. As an additional analysis, we also completed paired t-tests comparing the system with an entropy-based buffer and the condition without buffering based on entropy. These t-tests confirmed a statistically significant difference to support latency and throughput performance comparisons for the system that incorporated entropy, with $p < .01$. This indicated a level of confidence we could associate with both latency and throughput.

\textbf{Ablation Study:} An ablation study was performed in order to measure the isolated effect of the buffering algorithm based on entropy. The throughput of processing as measured in frames per second dropped by 18.2\% when the entropy module was switched off, confirming that the buffering component provided performance value. In addition to this, the experiments with varying entropy threshold parameters showed that tuning the thresholds improved the stability of processing by optimizing both responsiveness and computational workload.

\section{Applications and Ethical Considerations:}
The potential application of the proposed system encompasses a number of real-world surveillance applications (e.g. monitoring traffic patterns, perimeter monitoring of controlled access areas, or monitoring safety issues in industrial settings). The system has been designed with growing concerns around surveillance ethics and data privacy in mind and does not contain facial recognition functionality. The system was designed with a similar principles as the GDPR regulations by adopting privacy by design principles. This means that all recorded data is stored on-site in an encrypted format which serves to minimizes the potential impacts of a data breach or unintended access. A risk matrix that outlines the system's risk of false positive and false negative faults across several scenarios is included in the supplemental material to assist a risk aware deployment and assessment.

\section{Conclusion:} This study has introduced a new form of entropy-driven adaptive buffering framework that emphasizes performance improvements in real-time object detection for resource-constrained embedded devices. By implementing the buffering mechanism into the lightweight MobileNetV2 architecture, the proposed system achieves a good trade-off between detection accuracy, processing latency, and computational efficiency. As elaborated upon in our experimental evaluation and subsequent statistical validation and comparisons to other existing lightweight model alternatives, we believe the approach presented is performant. Future work includes establishing federated anomaly detection for privacy and scalability options, orchestrating the related edge-cloud architecture to support distributed processing, and devising an extended system to handle multi-camera inputs for a larger surveillance area and spatial reasoning.


\begin{thebibliography}{99}

\bibitem{guo2016entropy} Y. Guo et al., "Selecting Video Key Frames Based on Relative Entropy and the Extreme Studentized Deviate Test," \emph{IEEE Transactions on Multimedia}, 2016.

\bibitem{summarisation2024entropy} A. Sharma et al., "Summarisation of Surveillance Videos by Key-Frame Selection Using Local Foreground Entropy," \emph{Journal of Visual Communication and Image Representation}, 2024.

\bibitem{guo2015entropy} Y. Guo et al., "Relative Entropy and ESD for Keyframe Extraction in Video Sequences," \emph{Multimedia Tools and Applications}, 2015.

\bibitem{ijstr2019entropyframes} R. Das et al., "Frames Selection Based on Modified Entropy and Object Motion in Video Steganography," \emph{International Journal of Scientific \& Technology Research}, 2019.

\bibitem{entropy_thresholding2020} H. Kaur et al., "Entropy-Based Adaptive Thresholding for Video Edge Detection," \emph{International Journal of Image, Graphics and Signal Processing}, 2020.

\bibitem{wu2018adaframe} Z. Wu et al., "AdaFrame: Adaptive Frame Selection for Fast Video Recognition," \emph{CVPR}, 2018.

\bibitem{smart2020frames} Y. Wu et al., "SMART: Efficient Frame Selection for Video Action Recognition," \emph{ECCV}, 2020.

\bibitem{semantic2023compression} S. Cho et al., "Spatiotemporal Attention-Based Semantic Compression for Efficient Video Understanding," \emph{IEEE Transactions on Pattern Analysis and Machine Intelligence}, 2023.

\bibitem{ams2020streaming} P. Ma et al., "Real-Time Video Inference on Edge Devices via Adaptive Model Streaming," \emph{ACM SenSys}, 2020.

\bibitem{zhao2024edgesync} P. Zhao et al., "EdgeSync: Faster Edge-Model Updating via Adaptive Continuous Learning for Video Data Drift," \emph{arXiv preprint arXiv:2406.03811}, 2024.

\bibitem{edgevision2023} A. Wang et al., "EdgeVision: Distributed and Collaborative Video Analytics at the Edge," \emph{IEEE Transactions on Mobile Computing}, 2023.

\bibitem{mobilenetv2_2018} M. Sandler et al., "MobileNetV2: Inverted Residuals and Linear Bottlenecks," \emph{CVPR}, 2018.

\bibitem{lokhande2025mobilenetv2} H. Lokhande and S. R. Ganorkar, "Object Detection in Video Surveillance Using MobileNetV2 on Resource-Constrained Low-Power Edge Devices," \emph{Bulletin of Electrical Engineering and Informatics}, 2025.

\bibitem{confidenceaware2021} L. Chen et al., "Confidence-Aware Object Detection for Surveillance with Lightweight MobileNetV2 Backbone," \emph{Sensors}, 2021.

\bibitem{edge_frameworks2021} K. Singh et al., "Edge Computing-Based Surveillance Architectures: A Review," \emph{Computer Communications}, 2021.

\bibitem{triplemask2023entropy} J. Liu et al., "TripleMask: A Spatial Entropy-Aware Filter for Forensic Surveillance Analysis," \emph{Multimedia Tools and Applications}, 2023.

\bibitem{mittal2024survey} P. Mittal, "A Comprehensive Survey of Deep Learning-Based Lightweight Object Detection Models for Edge Devices," \emph{Artificial Intelligence Review}, 2024.

\end{thebibliography}
\end{document}